\documentclass[10pt,twocolumn,letterpaper]{article}

\usepackage{cvpr}
\usepackage{times}
\usepackage{epsfig}
\usepackage{graphicx}
\usepackage{amsmath}
\usepackage{amssymb}
\usepackage{xspace}
\usepackage{bm}

\usepackage[pagebackref=true,breaklinks=true,letterpaper=true,colorlinks,citecolor=blue,linkcolor=blue,bookmarks=false]{hyperref}

\cvprfinalcopy 

\def\cvprPaperID{326} 

\usepackage{subcaption}
\usepackage{xcolor}
\usepackage{rotating}
\usepackage[export]{adjustbox}
\makeatletter
\newcommand{\mycaption}[2]{\caption{\textbf{#1.}\xspace#2}}

\usepackage{booktabs}
\usepackage{array, makecell, tabularx}
\usepackage{multirow}

\def\model{SPLATNet\xspace}
\def\modelthree{SPLATNet$_{\text{3D}}$\xspace}
\def\modeljoint{SPLATNet$_{\text{2D-3D}}$\xspace}

\usepackage[normalem]{ulem}

\ifcvprfinal\fi
\begin{document}


\title{SPLATNet: Sparse Lattice Networks for Point Cloud Processing}

\author{Hang Su\\
UMass Amherst
\and
Varun Jampani\\
NVIDIA
\and
Deqing Sun\\
NVIDIA
\and
Subhransu Maji\\
UMass Amherst
\and
Evangelos Kalogerakis\\
UMass Amherst
\and
Ming-Hsuan Yang\\
UC Merced
\and
Jan Kautz\\
NVIDIA
}

\maketitle

\begin{abstract}
We present a network architecture for processing point clouds that directly operates on a collection of points represented as a sparse set of samples in a high-dimensional lattice.
Na\"ively applying convolutions on this lattice scales poorly, both in terms of memory and computational cost, as the size of the lattice increases.
Instead, our network uses sparse bilateral convolutional layers as building blocks.
These layers maintain efficiency by using indexing structures to apply convolutions only on occupied parts of the lattice, and allow flexible specifications of the lattice structure enabling hierarchical and spatially-aware feature learning, as well as joint 2D-3D reasoning. 
Both point-based and image-based representations can be easily incorporated in a network with such layers and the resulting model can be trained in an end-to-end manner.
We present results on 3D segmentation tasks where our approach outperforms existing state-of-the-art techniques.
\end{abstract}
\vspace{-3mm}

\section{Introduction}

Data obtained with modern 3D sensors such as laser scanners is predominantly in the \emph{irregular} format
of point clouds or meshes.
Analysis of point clouds has several useful applications such as robot
manipulation and autonomous driving. In this work, we aim to develop a new neural network architecture
for point cloud processing.

A point cloud consists of a \emph{sparse} and \emph{unordered} set of 3D points. 
These properties of point clouds make it difficult to use traditional convolutional neural network (CNN)
architectures for point cloud processing. As a result, existing approaches that directly operate on
point clouds are dominated by hand-crafted features. One way to use CNNs for point clouds is by
first pre-processing a given point cloud in a form that is amenable to standard spatial convolutions.
Following this route, most deep architectures for 3D point cloud analysis require pre-processing
of irregular point clouds into either voxel representations (\eg, ~\cite{wu2015shapenets,riegler2017octnet,wang2017ocnn}) or 
2D images by view projection (\eg, ~\cite{su15mvcnn,qi2016volmv,kalogerakis2017shapepfcn,cao2017sphericalprojection}).
This is due to the ease of implementing convolution operations on regular 2D or 3D grids.
However, transforming point cloud representation to either 2D images or 3D voxels would often result in artifacts and
more importantly, a loss in some natural invariances present in point clouds.

Recently, a few network architectures~\cite{qi2017pointnet,qi2017pointnetpp,zaheer2017deep} have been developed to directly work on point clouds.
One of the main drawbacks of these architectures is that they do not allow a flexible specification of
the extent of spatial connectivity across points (filter neighborhood). 
Both~\cite{qi2017pointnet} and~\cite{qi2017pointnetpp} use max-pooling to aggregate information across points either
globally~\cite{qi2017pointnet} or in a hierarchical manner~\cite{qi2017pointnetpp}.
This pooling aggregation may lose surface information because the spatial layouts of points are not explicitly considered.
It is desirable to capture spatial relationships in point clouds through more general convolution
operations while being able to specify filter extents in a flexible manner.

\begin{figure}
\begin{center}
\centerline{\includegraphics[width=1.0\columnwidth]{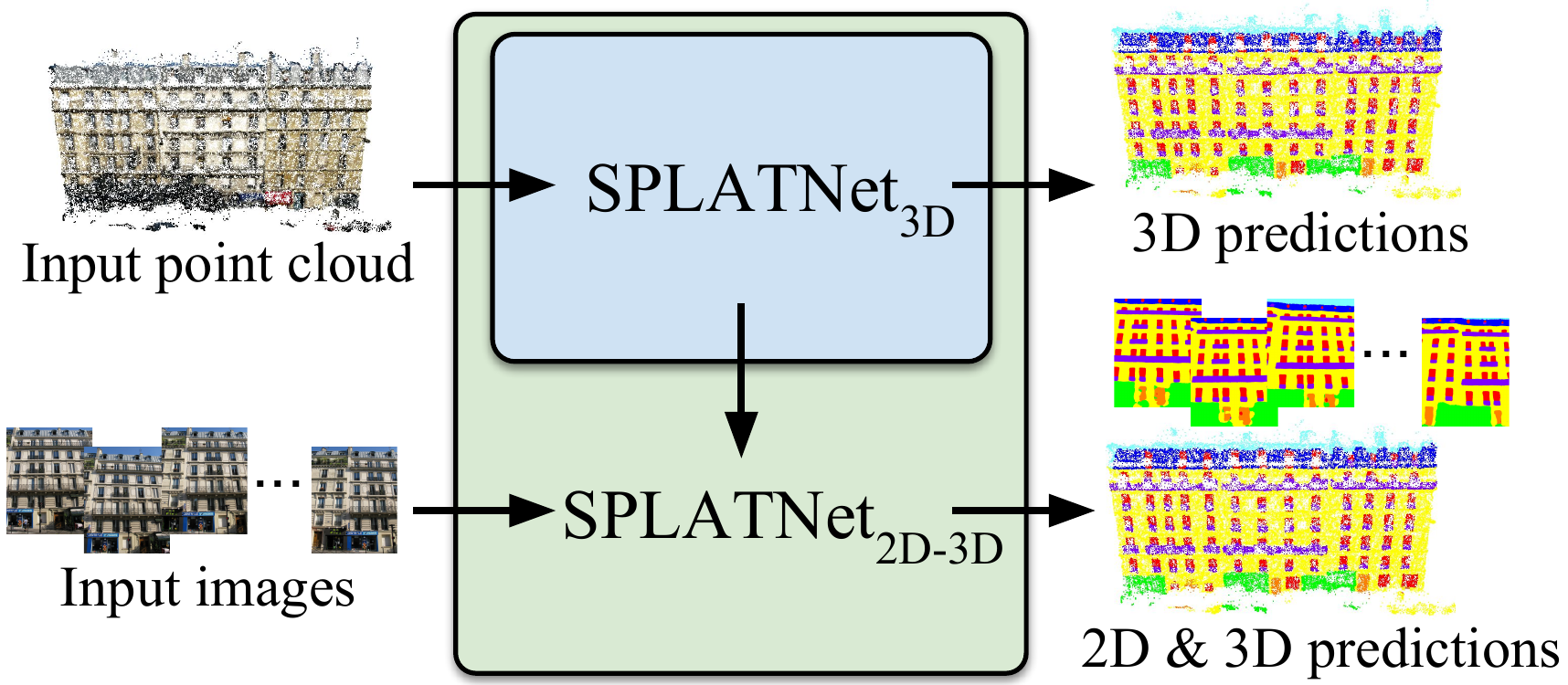}}
  \vspace{-0.2cm}
  \mycaption{From point clouds and images to semantics} {SPLATNet$_{\text{3D}}$ directly takes point cloud
  as input and predicts labels for each point. SPLATNet$_{\text{2D-3D}}$, on the other hand, jointly processes
  both point cloud and the corresponding multi-view images for better 2D and 3D predictions.}
  \label{fig:teaser}
\end{center}
\vspace{-0.8cm}
\end{figure}

In this work, we propose a generic and flexible neural network architecture for processing point clouds
that alleviates some of the aforementioned issues with existing deep architectures.
Our key observation is that the bilateral convolution layers (BCLs) proposed in~\cite{jampani2016learning,kiefel:iclr:2015} have several favorable properties
for point cloud processing. BCL provides a systematic way of filtering unordered points while enabling 
flexible specifications of the underlying lattice structure
on which the convolution operates.
BCL smoothly maps input points onto a sparse lattice, performs 
convolutions on the sparse lattice
and then smoothly interpolates the filtered signal back onto the original input points.
With BCLs as building blocks, we propose a new neural network architecture,
which we refer to as \model~(SParse LATtice Networks), that does hierarchical and spatially-aware feature learning for unordered points.
\model has several advantages for point cloud processing: 
\vspace{-1mm}
\begin{itemize}
    \setlength\itemsep{0em}
    \item \model~takes the point cloud as input and does not require any pre-processing to voxels or images.
    \item \model~allows an easy specification of filter neighborhood as in standard CNN architectures.
    \item With the use of hash table, our network can efficiently deal with sparsity in the input point cloud by convolving  only at locations where data is present.
    \item SPLATNet computes hierarchical and spatially-aware features of an input point cloud with sparse
    and efficient lattice filters.
    \item In addition, our network architecture allows an easy mapping of 2D points into 3D space and vice-versa. Following this,
    we propose a joint 2D-3D deep architecture that processes both the multi-view 2D images and the corresponding 3D point cloud in a single forward pass while being end-to-end learnable.
\end{itemize}
\vspace{-1mm}

The inputs and outputs of two versions of the proposed network, SPLATNet$_{\text{3D}}$ and SPLATNet$_{\text{2D-3D}}$, 
are depicted in Figure~\ref{fig:teaser}.
We demonstrate the above advantages with experiments on point cloud segmentation. 
Experiments on both RueMonge2014 facade segmentation~\cite{riemenschneider2014learning} and ShapeNet part segmentation~\cite{yi2016scalable} 
demonstrate the superior performance of our technique compared to state-of-the-art techniques, 
while being computationally efficient.

\section{Related Work}
\label{sec:related}

Below we briefly review existing deep learning approaches for 3D shape processing and explain differences with our work.

\vspace{-0.35cm}
\paragraph{Multi-view and voxel networks.} Multi-view networks pre-process shapes into a set of 2D rendered images encoding surface depth and normals under various 2D projections~\cite{su15mvcnn,qi2016volmv,bai2016gift,kalogerakis2017shapepfcn,cao2017sphericalprojection,huang2018lmvcnn}. These networks take advantage of high resolution in the input rendered images and transfer learning through fine-tuning of 2D pre-trained image-based architectures. On the other hand, 2D projections can cause surface information loss due to self-occlusions, while viewpoint selection is often performed through heuristics that are not necessarily optimal for a given task.

Voxel-based methods convert the input 3D shape representation into a 3D volumetric grid. Early voxel-based architectures executed convolution in regular, fixed voxel grids, and were limited to low shape resolutions due to high memory and computation costs~\cite{wu2015shapenets,maturana2015voxnets,qi2016volmv,brock2016anothervoxnet,garciagarcia2016pointnetwhichisvoxnetactually,sedaghat2017orion}. Instead of using fixed grids, more recent approaches pre-process the input shapes into adaptively subdivided, hierarchical grids with denser cells placed near the surface ~\cite{riegler2017octnet,riegler2017OctNetFusion,klokov2017escape,wang2017ocnn,tatarchenko2017octree}. As a result, they have much lower computational and memory overhead.  On the other hand, convolutions are often still executed away from the surface, where most of the shape information resides. An alternative approach is to constrain the execution of volumetric convolutions only along the input sparse set of active voxels of the grid~\cite{graham2017ssnet}. 
Our approach generalizes this idea to high-dimensional permutohedral lattice convolutions.
In contrast to previous work, we do not require pre-processing points into voxels that may cause discretization artifacts and surface information loss. We smoothly map the input surface signal to our sparse lattice, perform convolutions over this lattice, and smoothly interpolate the filter responses back to the input surface. In addition, our architecture can easily incorporate feature representations originating from both 3D point clouds and rendered images within the same lattice, getting the best of both worlds. 

\vspace{-0.35cm}
\paragraph{Point cloud networks.} Qi \etal~\cite{qi2017pointnet} pioneered another type of deep networks having the advantage of directly operating on point clouds. The networks learn spatial feature representations for each input point, then the point features are aggregated across the whole point set~\cite{qi2017pointnet}, or hierarchical surface regions~\cite{qi2017pointnetpp} through max-pooling. This aggregation may lose surface information since the spatial layout of points is not explicitly considered. In our case, the input points are mapped to a sparse lattice where convolution can be efficiently formulated and spatial relationships in the input data can be effectively captured through flexible filters. 

\vspace{-0.35cm}
\paragraph{Non-Euclidean networks.} An alternative approach is to represent the input surface as a graph (\eg, a polygon mesh or point-based connectivity graph), convert the graph into its spectral representation, then perform convolution in the spectral domain~\cite{Bruna2013spectral,Henaff2015spectral,Defferrard2016spectral,Boscaini2015spectral}. 
However, structurally different shapes tend to have largely different spectral bases, and thus lead to poor generalization. 
Yi \etal~\cite{yi2017syncspeccnn} proposed aligning shape basis functions through a spectral transformer, which, however, requires a robust initialization scheme.
 Another class of methods embeds the input shapes into 2D parametric domains and then execute convolutions within these domains~\cite{Sinha2016,Maron2017CNN,Ezuz2017}. However, these embeddings can suffer from spatial distortions or require topologically consistent input shapes. Other methods parameterize the surface into local patches and execute surface-based convolution within these patches~\cite{masci2015geodesic,Boscaini2016,Monti2017}. Such non-Euclidean networks have the advantage of being invariant to surface deformations, yet this invariance might not always be desirable in man-made object segmentation and classification tasks where large deformations may change the underlying shape or part functionalities and semantics. We refer to Bronstein \etal~\cite{bronstein2017geometric} for an excellent review of spectral, patch- and graph-based methods.

\vspace{-0.35cm}
\paragraph{Joint 2D-3D networks.} FusionNet~\cite{hedge2016fusionnet} combines shape classification scores from a volumetric and a multi-view network, yet this fusion happens at a late stage, after the final fully connected layer of these networks, and does not jointly consider their intermediate local and global feature representations. In our case, the 2D and 3D feature representations are mapped onto the same lattice, enabling end-to-end learning from both types of input representations.


\section{Bilateral Convolution Layer}
\label{sec:review_bcl}

In this section, we briefly review the Bilateral Convolution Layer (BCL) 
that forms the basic building block of our SPLATNet architecture for point clouds.
BCL provides a way to incorporate sparse high-dimensional
filtering inside neural networks. In~\cite{jampani2016learning,kiefel:iclr:2015}, BCL was proposed as a learnable generalization
of bilateral filtering~\cite{tomasi1998bilateral,aurich1995non}, 
hence the name `Bilateral Convolution Layer'.
Bilateral filtering involves a projection of a given 2D image into
a higher-dimensional space (\eg, space defined by position and color) and
is traditionally limited to hand-designed filter kernels. BCL provides a way to learn filter
kernels in high-dimensional spaces for bilateral filtering.
BCL is also shown to be useful for information propagation across video 
frames~\cite{jampani2017video}.
We observe that BCL has several favorable properties to
filter data that is inherently sparse and high-dimensional, like point clouds.
Here, we briefly describe how a BCL works and then discuss its properties.

\begin{figure}
\vspace{-0.2cm}
\begin{center}
\centerline{\includegraphics[width=1.1\columnwidth]{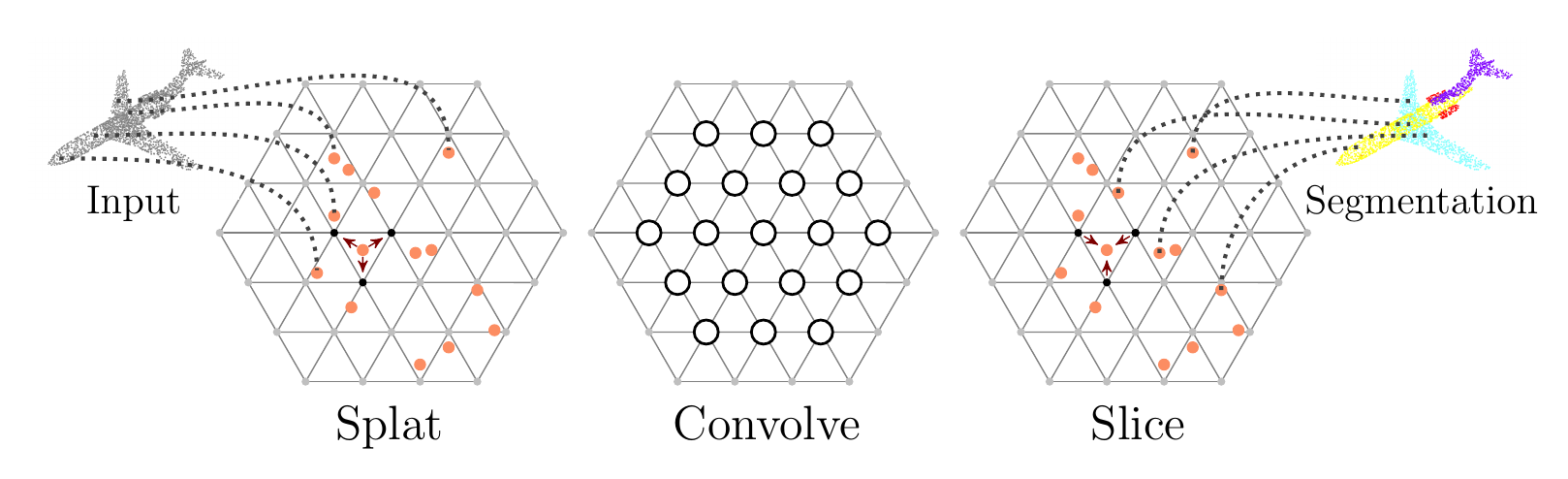}}
  \vspace{-0.2cm}
  \mycaption{Bilateral Convolution Layer} {\emph{Splat}: BCL first interpolates input features $F$ onto a $d_l$-dimensional permutohedral lattice defined by the lattice features $L$ at input points. \emph{Convolve}: BCL then does $d_l$-dimensional convolution over this sparsely populated lattice. \emph{Slice}: The filtered signal is then interpolated back onto the input signal. For illustration, input and output are shown as point cloud and the corresponding segmentation labels.}
  \label{fig:bcl}
\end{center}
\vspace{-.8cm}
\end{figure}

\subsection{Inputs to BCL} 
Let $F \in \mathbb{R}^{n \times d_f}$ be the given \emph{input features} to a BCL, where $n$ denotes
the number of input points and $d_f$ denotes the dimensionality of input features at each point.
For 3D point clouds, input features can be low-level features such as color, position, \etc, and can also
be high-level features such as features generated by a neural network.

One of the interesting characteristics of BCL is that it allows a flexible specification
of the lattice space in which the convolution operates. This is specified as \emph{lattice features} at 
each input point. Let $L \in \mathbb{R}^{n \times d_l}$ denote lattice features at input points
with $d_l$ denoting the dimensionality of the feature space in which convolution operates.
For instance, the lattice features can be 
point position and color ($XYZRGB$)
that define a 6-dimensional filtering space for BCL. For standard 3D spatial filtering of point clouds,
$L$ is given as the position ($XYZ$) of each point.
Thus BCL takes input features $F$ and lattice features $L$ of input points and performs $d_l$-dimensional filtering of the points. 

\begin{figure*}[!ht]
    \vspace{-0.2cm}
    \centering
    \includegraphics[width=0.95\textwidth]{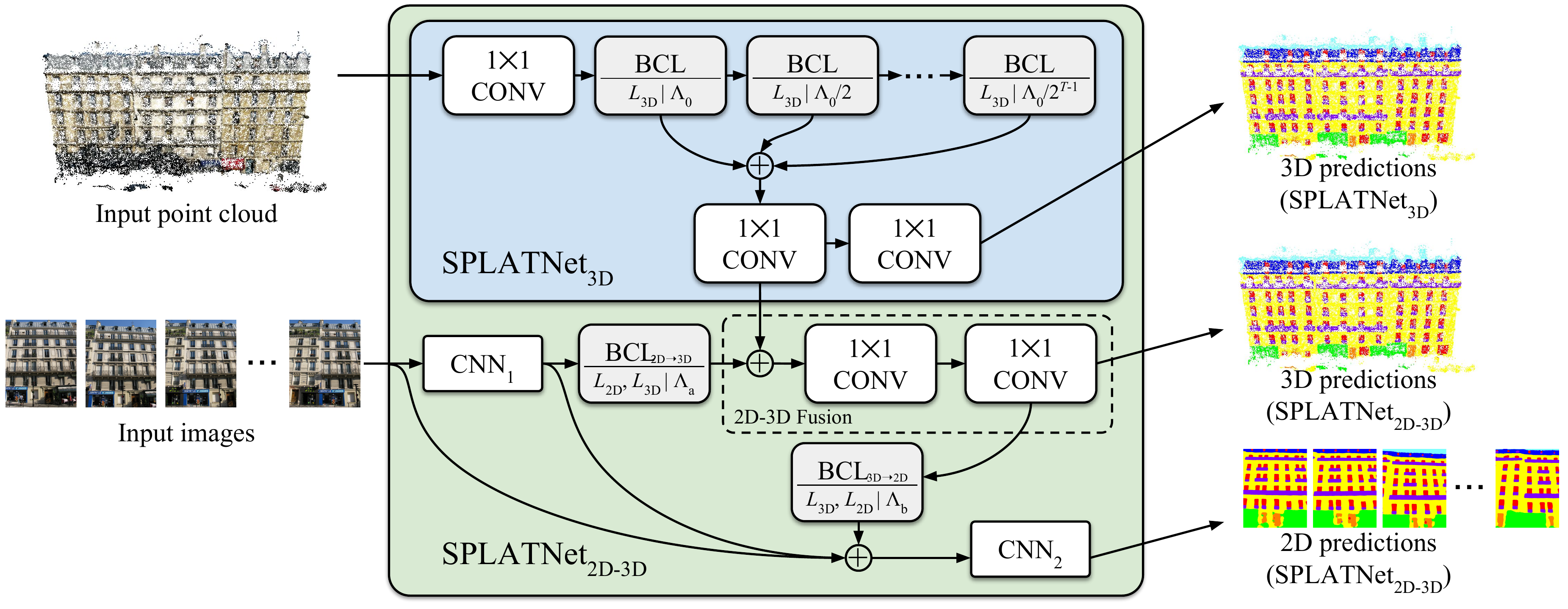}
    \vspace{-0.2cm}
    \mycaption{SPLATNet}{Illustration of inputs, outputs and network architectures for SPLATNet$_{\text{3D}}$ and SPLATNet$_{\text{2D-3D}}$.}
    \label{fig:splatnet}
    \vspace{-0.25cm}
\end{figure*}

\subsection{Processing steps in BCL}
As illustrated in Figure~\ref{fig:bcl}, 
BCL has three processing steps, \emph{splat}, \emph{convolve} and \emph{slice}, that work as follows.

\vspace{-0.35cm}
\paragraph{Splat.} BCL first projects the input features $F$ onto the $d_l$-dimensional
lattice defined by the lattice features $L$, via barycentric interpolation. 
Following~\cite{adams2010fast}, BCL uses a permutohedral
lattice instead of a standard Euclidean grid for efficiency purposes. The size of lattice simplices
or space between the grid points is controlled by scaling the lattice features $\Lambda L$,
where $\Lambda$ is a diagonal $d_l\times d_l$ scaling matrix.

\vspace{-0.35cm}
\paragraph{Convolve.} Once the input points are projected onto the $d_l$-dimensional lattice,
BCL performs $d_l$-dimensional convolution on the splatted signal with learnable filter kernels.
Just like in standard spatial CNNs, BCL allows an easy specification of filter neighborhood in the $d_l$-dimensional space.

\vspace{-0.35cm}
\paragraph{Slice.} The filtered signal is then mapped back to the input points via barycentric
interpolation. The resulting signal can be passed on to other BCLs for further processing. 
This step is called `slicing'. BCL allows slicing the filtered signal onto a different set
of points other than the input points. 
This is achieved by specifying a different
set of lattice features $L^{out}\in \mathbb{R}^{m \times d_l}$ at $m$ output points of interest.

All the above three processing steps in BCL can be written as matrix multiplications:
\vspace{-0.2cm}
\begin{equation}
\vspace{-0.2cm}
    \hat{F}_c = S_{slice} B_{conv} S_{splat} F_c,
\end{equation}
where $F_c$ denotes the $c^{th}$ column/channel of the input feature $F$ and $\hat{F}_c$ denotes
the corresponding filtered signal.

\subsection{Properties of BCL}

There are several properties of BCL that makes it particularly convenient for point cloud
processing. Here, we mention some of those properties:
\vspace{-1mm}
\begin{itemize}
    \setlength\itemsep{0mm}
    \item The input points to BCL need not be ordered or lie on a grid as
    they are projected onto a $d_l$-dimensional grid defined by lattice features $L^{in}$.
    \item The input and output points can be different for BCL with 
    the specification of different input and output lattice features $L^{in}$ and $L^{out}$.
    \item Since BCL allows separate specifications of input and lattice
    features, input signals can be projected into a different dimensional space for filtering.
    For instance, a 2D image can be projected into 3D space for filtering.
    \item Just like in standard spatial convolutions, BCL allows an easy specification
    of filter neighborhood. 
    \item Since a signal is usually sparse in high-dimension, BCL uses
    hash tables to index the populated vertices and does convolutions only at those locations.
    This helps in efficient processing of sparse inputs.
\end{itemize} 
\vspace{-1mm}

Refer to~\cite{adams2010fast} for more information about sparse high-dimensional Gaussian filtering
on a permutohedral lattice and refer to~\cite{jampani2016learning} for more details on BCL.

\section{SPLATNet$_{\text{3D}}$ for Point Cloud Processing}
\label{sec:bpcn_3d}

We first introduce SPLATNet$_{\text{3D}}$, an instantiation of our proposed network architecture which operates directly on 3D point clouds and is readily applicable to many important 3D tasks. 
The input to SPLATNet$_{\text{3D}}$ is a 3D point cloud $P \in \mathbb{R}^{n \times d}$, where $n$ denotes the number of points and $d \ge 3$ denotes the number of feature dimensions including point locations $XYZ$. 
Additional features are often available either directly from 3D sensors 
or through pre-processing. These can be RGB color, surface normal, curvature, \etc. at the input points.
Note that input features $F$ of the first BCL and lattice features $L$ in the network each 
comprises a subset of the $d$ feature dimensions: $d_f\le d, d_l \le d$.

As output, SPLATNet$_{\text{3D}}$ produces per-point predictions. Tasks like 3D semantic segmentation and 
3D object part labeling fit naturally under this framework. 
With simple techniques such as global pooling \cite{qi2017pointnet}, SPLATNet$_{\text{3D}}$ can be modified to 
produce a single output vector and thus can be extended to other tasks such as classification.

\vspace{-0.35cm}
\paragraph{Network architecture.}
The architecture of SPLATNet$_{\text{3D}}$ is depicted in Figure~\ref{fig:splatnet}. The network starts with a single $1\times1$ CONV layer followed by a series of BCLs. The $1\times1$ CONV layer processes
each input point separately without any data aggregation. The functionality of BCLs is already explained
in Section~\ref{sec:review_bcl}. For SPLATNet$_{\text{3D}}$, we use $T$ BCLs each operating on
a 3D lattice ($d_l=3$) constructed using 
3D point locations $XYZ$ as lattice features,
$L^{in} = L^{out} \in \mathbb{R}^{n \times 3}$. 
We note that different BCLs can use different lattice scales $\Lambda$.
Recall from Section~\ref{sec:review_bcl} that $\Lambda$ is a diagonal matrix that controls the spacing between the grid points in the lattice. For BCLs in \modelthree, we use the same lattice scales along each of the $X$, $Y$ and $Z$ directions, \ie, $\Lambda = \lambda I_3$, where $\lambda$ is a scalar and $I_3$ denotes a $3\times 3$ identity matrix.
We start with an initial lattice scale $\lambda_0$ for the first BCL and subsequently divide the lattice scale by a factor of 2 ($\lambda_t = \lambda_{t-1} / 2$) 
for the next $T-1$ BCLs. 
In other words, \modelthree with $T$ BCLs use the following lattice scales: $(\Lambda_0, \Lambda_0/2, \dots, \Lambda_0/2^{T-1})$.
Lower lattice scales imply coarser lattices and larger receptive fields for the filters. 
Thus, in SPLATNet$_{\text{3D}}$, deeper BCLs have longer-range connectivity between input points compared 
to earlier layers. We will discuss more about the effects of different lattice spaces and their scales later.
Like in standard CNNs, SPLATNet allows an easy specification of filter neighborhoods.
For all the BCLs, we use filters operating on 
one-ring neighborhoods
and refer to the supp. material
for details on the number of filters per layer.

The responses of the $T$ BCLs are concatenated and then passed through two additional $1\times1$ CONV layers. 
Finally, a softmax layer produces point-wise class label probabilities.
The concatenation operation aggregates information from BCLs operating at different lattice scales. 
Similar techniques of concatenating outputs from network layers at different depths
have been useful in 2D CNNs~\cite{hariharan2015hypercolumns}.
All parameterized layers, except for the last CONV layer, are followed by ReLU and BatchNorm.
More details about the network architecture are given in the supp. material.

\vspace{-0.35cm}
\paragraph{Lattice spaces and their scales.}
The use of BCLs in SPLATNet allows easy specifications of lattice spaces via lattice features and also lattice scales via a scaling matrix.

\begin{figure}[!htb]
\vspace{1.5mm}
\centering
\captionsetup[subfigure]{labelformat=empty}
\begin{subfigure}{.155\textwidth}
\centering
\includegraphics[height=3.5cm]{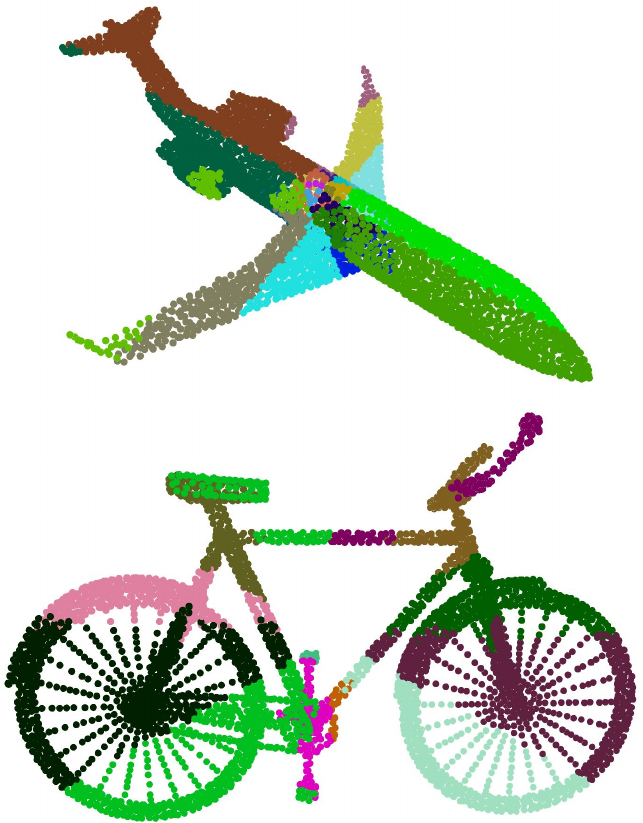}
\caption{$(x,y,z), I_3$}
\end{subfigure}
\begin{subfigure}{.155\textwidth}
\centering
\includegraphics[height=3.5cm]{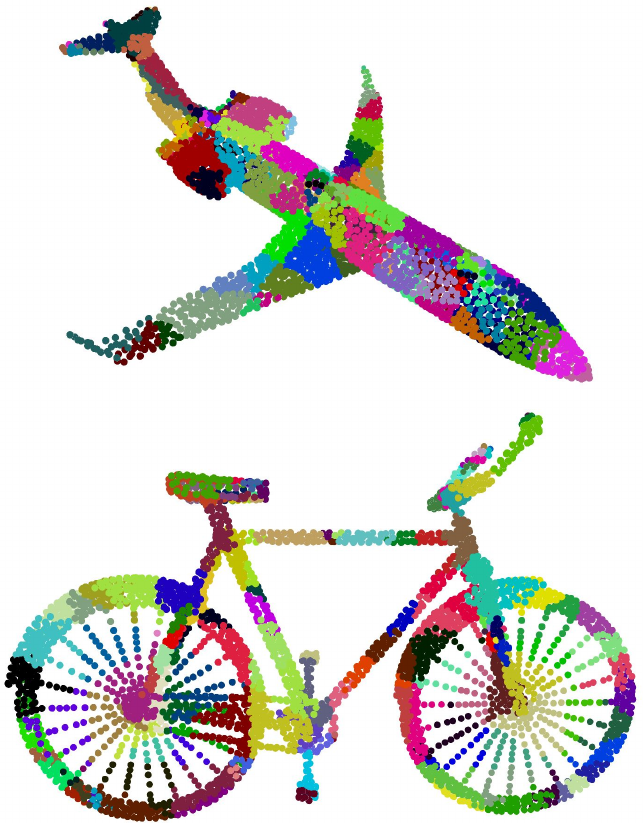}
\caption{$(x,y,z), 8 I_3$}
\end{subfigure}
\begin{subfigure}{.155\textwidth}
\centering
\includegraphics[height=3.5cm]{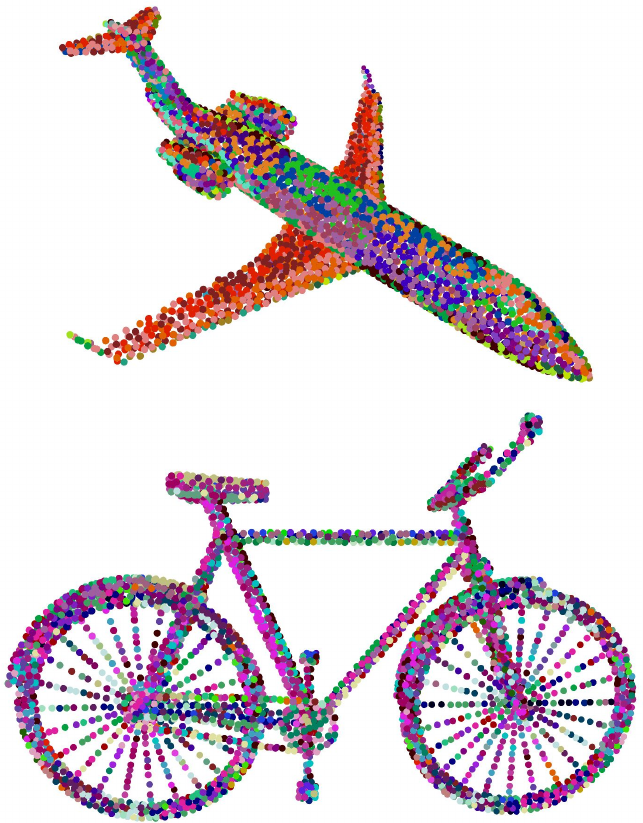}
\caption{$(n_x,n_y,n_z), I_3$}
\end{subfigure}
\vspace{-0.1cm}
\mycaption{Effect of different lattice spaces and scales}{Visualizations for different
lattice feature spaces $L = (x,y,z), (x,y,z), (n_x,n_y,n_z)$ along with lattice scales
$\Lambda = I_3, 8I_3, I_3$. $(n_x,n_y,n_z)$ refers to point normals.
All points falling in the same lattice cell are colored the same.}
\label{fig:lattice_viz}
\vspace{-0.3cm}
\end{figure}

Changing the lattice scales $\Lambda$ directly affects the resolution of the signal on which the convolution operates. This gives us direct control over the receptive fields of network layers.
Figure~\ref{fig:lattice_viz} shows lattice cell visualizations for different lattice spaces and scales.
Using coarser lattice can increase the effective receptive field of a filter.
Another way to increase the receptive field of a filter is by increasing its neighborhood size. 
But, in high-dimensions, this will significantly increase the number of filter parameters. For instance, 3D filters of size $3, 5, 7$ on a regular Euclidean grid have $3^3=27, 5^3=125, 7^3=343$ parameters respectively.
On the other hand, 
making the lattice coarser would not increase the number of filter parameters leading to more computationally efficient network architectures.

We observe that it is beneficial to use finer lattices (larger lattice scales) earlier in the network, and then coarser lattices (smaller lattice scales) going deeper. This is consistent with the common knowledge in 2D CNNs: increasing receptive field gradually through the network can help build hierarchical representations with varying spatial extents and abstraction levels. 

Although we mainly experiment with $XYZ$ lattices in this work, BCL allows for other lattice spaces such as position and color space ($XYZRGB$) or normal space. Using different lattice spaces enforces different connectivity across input points that may be beneficial to the task. In one of the experiments, we experimented with a variant of SPLATNet$_{\text{3D}}$, where
we add an extra BCL with position and normal lattice features ($XYZn_xn_yn_z$) and observed minor
performance improvements.

\section{Joint 2D-3D Processing with SPLATNet$_{\text{2D-3D}}$}
\label{sec:bpcn_2d3d}

Oftentimes, 3D point clouds are accompanied by 2D images of the same target. For instance, many modern 3D
sensors capture RGBD streams and perform 3D reconstruction to obtain 3D point clouds, resulting
in both 2D images and point clouds of a scene together with point correspondences between 2D and 3D.
One could also easily sample point clouds along with 2D renderings from a given 3D mesh.
When such aligned 2D-3D data is present, SPLATNet provides an extremely flexible framework for joint 
processing. We propose SPLATNet$_{\text{2D-3D}}$, another SPLATNet instantiation designed for such joint
processing.

The network architecture of the SPLATNet$_{\text{2D-3D}}$ is depicted in the green box of Figure~\ref{fig:splatnet}.
SPLATNet$_{\text{2D-3D}}$ encompasses SPLATNet$_{\text{3D}}$ as one of its components and adds extra computational 
modules for joint 2D-3D processing. Next, we explain each of these extra components of SPLATNet$_{\text{2D-3D}}$,
in the order of their computations.



\vspace{-0.35cm}
\paragraph{CNN$_1$.} First, we process the given multi-view 2D images using 
a 2D segmentation CNN, which we refer to as CNN$_1$.
In our experiments, we use the DeepLab~\cite{chen2014semantic} architecture
for CNN$_1$ and initialize the network weights with those pre-trained on
PASCAL VOC segmentation~\cite{Everingham15}.

\begin{figure}[!t]
\vspace{-0.5cm}
\begin{center}
\centerline{\includegraphics[width=1.05\columnwidth]{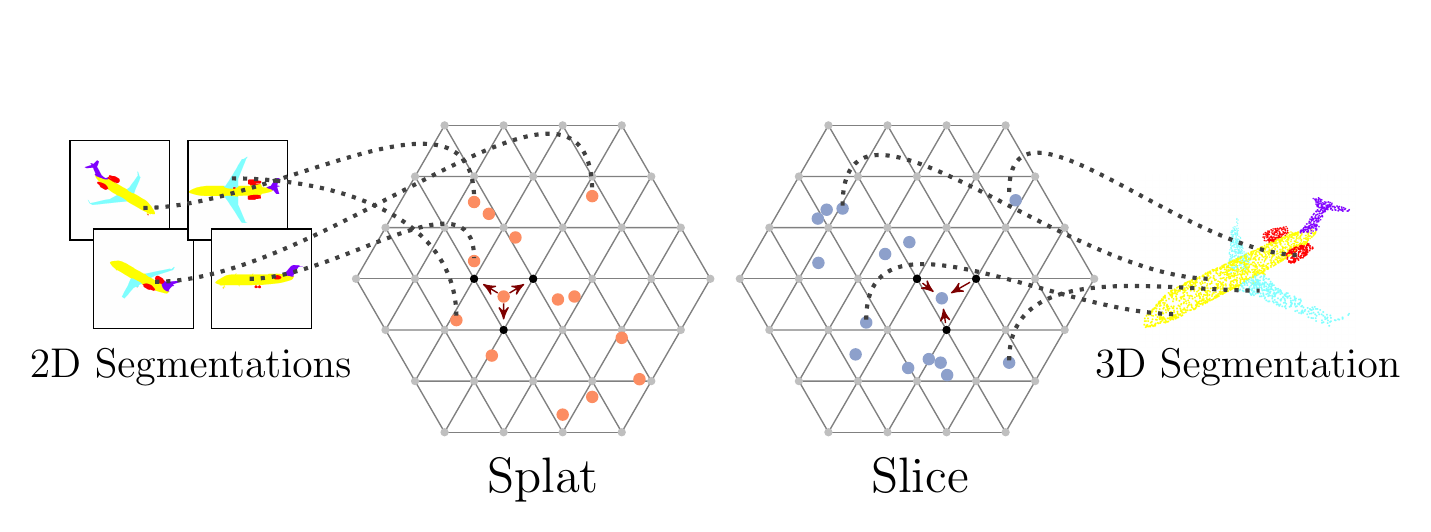}}
  \vspace{-0.2cm}
  \mycaption{2D to 3D projection} {Illustration of 2D to 3D projection using \emph{splat} and \emph{slice}usingsplatandsliceoperations. Given input features of 2D images, pixels are projected onto a 3D permutohedral lattice defined by 3D positional lattice features. The splatted signal
  is then sliced onto the points of interest in a 3D point cloud.}
    \label{fig:map_2d_3d}
\end{center}
\vspace{-0.7cm}
\end{figure}

\vspace{-0.35cm}
\paragraph{BCL$_{\text{2D}\rightarrow\text{3D}}$.} 
CNN$_1$ outputs features of the image pixels, 
whose 3D locations often do not exactly correspond to points in the 3D point cloud. We project 
information from the pixels onto the point cloud using a BCL with only \emph{splat} and \emph{slice} operations. 
As mentioned in Section~\ref{sec:review_bcl}, one of the interesting properties of BCL is that it allows for different
input and output points by separate specifications of input and output lattice features, $L^{in}$ and $L^{out}$.
Using this property, we use BCL to \emph{splat} 2D features onto the 3D lattice space and then \emph{slice} the 3D splatted signal on the point cloud.
We refer to this BCL, without a convolution operation, as BCL$_{\text{2D}\rightarrow\text{3D}}$ 
as illustrated in Figure~\ref{fig:map_2d_3d}. Specifically, we use 3D locations of the image pixels
as input lattice features, $L^{in}=L_{2D} \in \mathbb{R}^{m \times 3}$, where $m$ denotes the number
of input image pixels. In addition, we use 3D locations of points in the point cloud as output lattice
features, $L^{out}=L_{3D} \in \mathbb{R}^{n \times 3}$, which are the same lattice features used in SPLATNet$_{\text{3D}}$.
The lattice scale, $\Lambda_a$, controls the smoothness of the projection and can be adjusted
according to the sparsity of the point cloud.

\vspace{-0.35cm}
\paragraph{2D-3D Fusion.} At this point, we have the result of CNN$_1$ projected onto 3D points and also
the intermediate features from SPLATNet$_{\text{3D}}$ that exclusively operates on the input point cloud. Since both
of these signals are embedded in the same 3D space, we concatenate these two signals and then use a series of $1\times1$ CONV layers for further processing. The output of the `2D-3D Fusion' module is 
passed on to a softmax layer to compute class probabilities at each input point of the point cloud.


\vspace{-0.35cm}
\paragraph{BCL$_{\text{3D}\rightarrow\text{2D}}$.} Sometimes, we are also interested in segmenting 2D images and
want to leverage relevant 3D information for better 2D segmentation. For this purpose, we back-project
the 3D features computed by the `2D-3D Fusion' module onto the 2D images by a BCL$_{\text{2D}\rightarrow\text{3D}}$ module. This is the reverse operation of BCL$_{\text{2D}\rightarrow\text{3D}}$, where the
input and output lattice features are swapped. Similarly, a hyper-parameter $\Lambda_b$ controls the smoothness of the projection.


\vspace{-0.35cm}
\paragraph{CNN$_2$.} We then concatenate the output from CNN$_1$, input images and the output
of BCL$_{\text{3D}\rightarrow\text{2D}}$, and pass them through another 2D CNN, CNN$_2$, to obtain refined
2D semantic predictions. In our experiments, we find that a simple 2-layered network is good enough
for this purpose.


All components in this 2D-3D joint processing framework are differentiable, and can be trained end-to-end. Depending on the availability of 2D or 3D ground-truth labels, loss functions can be defined on either one of the two domains, or on both domains in a multi-task learning setting. More details of the network architecture
are provided in the supp. material. We believe that this joint processing capability offered by SPLATNet$_{\text{2D-3D}}$ can result in better predictions for both 2D images and 3D point clouds. For 2D images, leveraging 3D features helps in view-consistent predictions across multiple viewpoints. 
For point clouds, incorporating 2D CNNs help leverage powerful 2D deep CNN features computed on high-resolution images.

\section{Experiments}
\label{sec:exp}

We evaluate \model on tasks on two different benchmark datasets of
RueMonge2014~\cite{riemenschneider2014learning} and ShapeNet~\cite{yi2016scalable}.
On RueMonge2014, we conducted experiments on the tasks of 3D point cloud labeling
and multi-view image labeling. On ShapeNet, we evaluated \model on 3D part segmentation. We use Caffe~\cite{jia2014caffe} neural network framework for all the experiments.
Full code and trained models are publicly available on our project website\footnote{\url{http://vis-www.cs.umass.edu/splatnet}}.

\subsection{RueMonge2014 facade segmentation}
\label{sec:exp_facade}

\begin{table}[!tb]
    \scriptsize
    \mycaption{Results on facade segmentation}{Average IoU scores and approximate runtimes
    for point cloud labeling and 2D image labeling using different techniques. Runtimes indicate the time taken to segment the entire test data (202 images sequentially for 2D and a point cloud for 3D).}
    \centering
    \small
    \begin{tabular}{p{3.4cm}>{\centering\arraybackslash}p{1.7cm}>{\centering\arraybackslash}p{2.0cm}}
        \toprule
        Method & Average IoU & Runtime (min) \\
        \midrule
        \multicolumn{2}{l}{\textit{With only 3D data}} & \\
        OctNet~\cite{riegler2017octnet}              &  59.2  & - \\
        Autocontext$_{\text{3D}}$~\cite{gadde2017efficient}     &  54.4  & 16 \\
        \modelthree (Ours)      & \textbf{65.4}  & 0.06 \\
        \midrule
        \multicolumn{2}{l}{\textit{With both 2D and 3D data}} &  \\
        Autocontext$_{\text{2D-3D}}$~\cite{gadde2017efficient}   &  62.9  &  87\\
        \modeljoint (Ours) & \textbf{69.8} & 1.20 \\
        \bottomrule
    \end{tabular}
    \subcaption{Point cloud labeling}
    \label{tab:facade_results_3d}
    \begin{tabular}{p{3.4cm}>{\centering\arraybackslash}p{1.7cm}>{\centering\arraybackslash}p{2.0cm}}
        \toprule
        Method & Average IoU & Runtime (min) \\
        \midrule
        Autocontext$_{\text{2D}}$~\cite{gadde2017efficient}     &  60.5  & 117 \\
        Autocontext$_{\text{2D-3D}}$~\cite{gadde2017efficient}     &  62.7  & 146\\
        DeepLab$_{\text{2D}}$~\cite{chen2014semantic} & 69.3 & 0.84\\
        \modeljoint (Ours)   & \textbf{70.6} &  4.34 \\
        \bottomrule
    \end{tabular}
    \subcaption{Multi-view image labeling}
    \label{tab:facade_results_2d}
    \vspace{-3mm}
\end{table}

\begin{figure*}[!htb]
\captionsetup[subfigure]{labelformat=empty}
\begin{subfigure}{.24\textwidth}
\centering
\includegraphics[height=3.5cm]{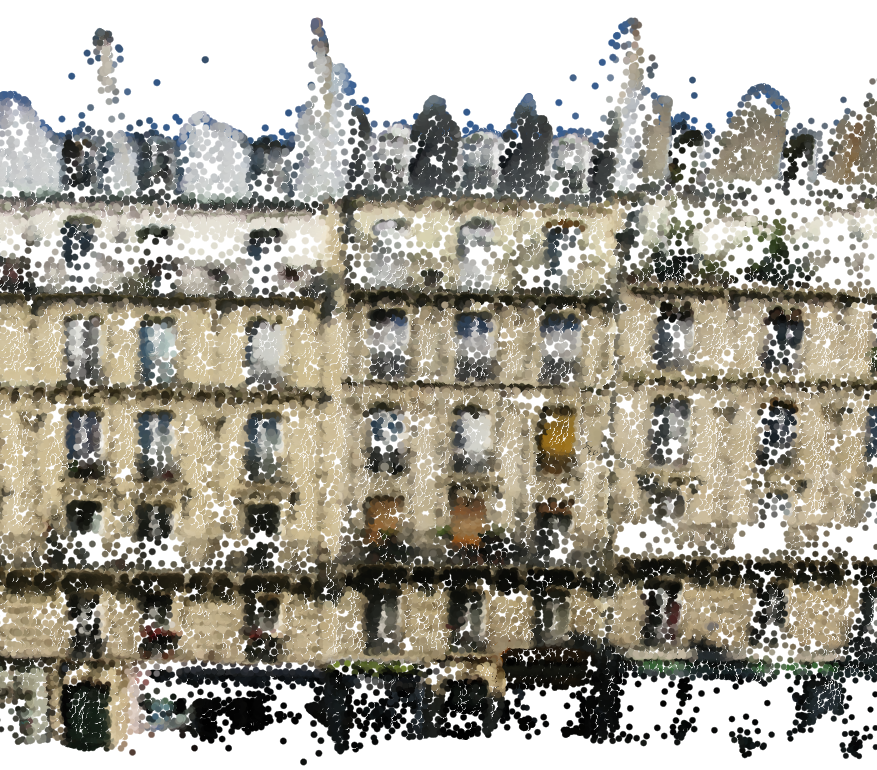}
\caption{Input Point Cloud}
\end{subfigure}
\begin{subfigure}{.24\textwidth}
\centering
\includegraphics[height=3.5cm]{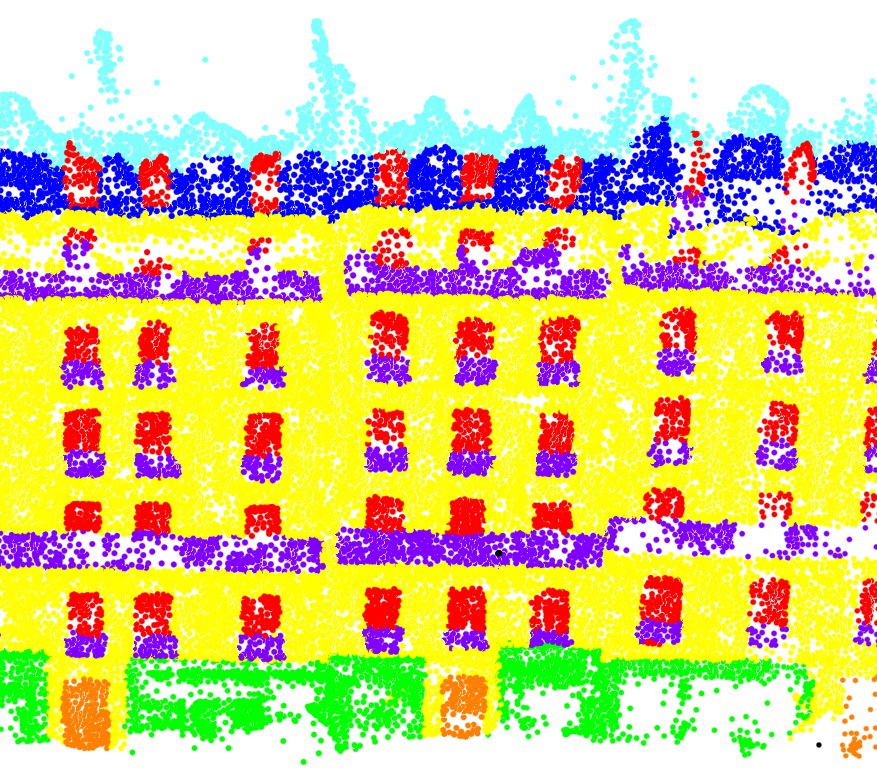}
\caption{Ground truth}
\end{subfigure}
\begin{subfigure}{.24\textwidth}
\centering
\includegraphics[height=3.5cm]{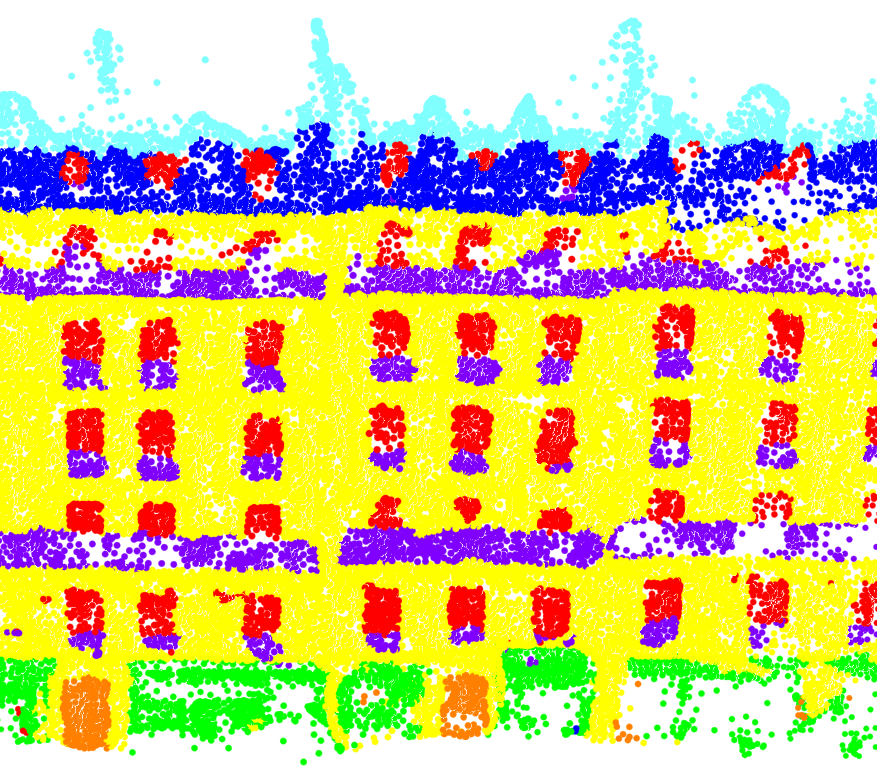}
\caption{\modelthree}
\end{subfigure}
\begin{subfigure}{.24\textwidth}
\centering
\includegraphics[height=3.5cm]{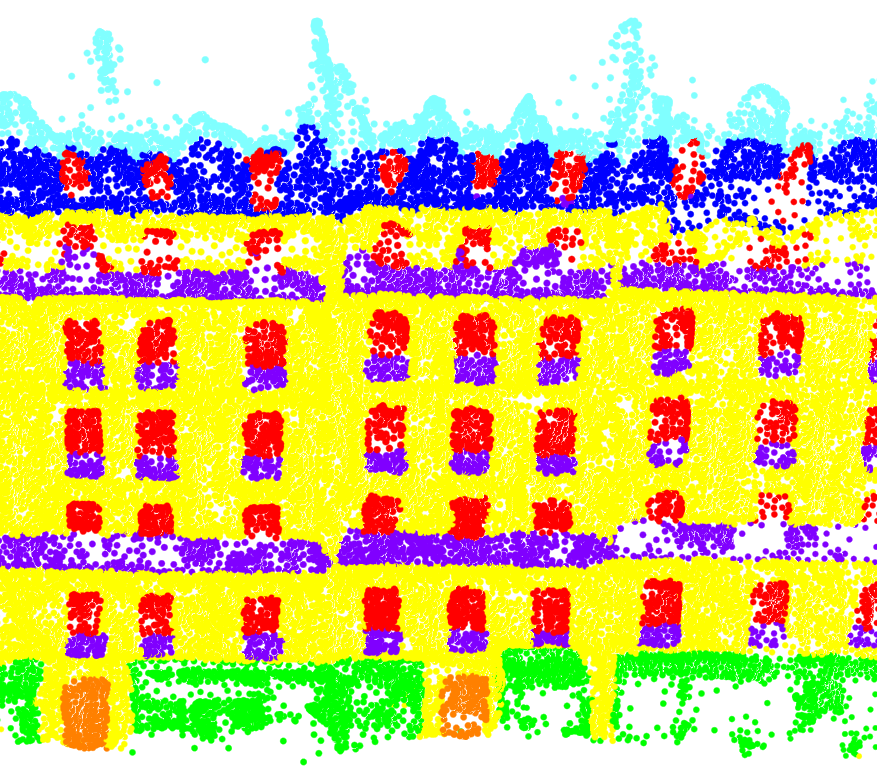}
\caption{\modeljoint}
\end{subfigure}
\vspace{-.1cm}
\mycaption{Facade point cloud labeling}{Sample visual results of \modelthree and \modeljoint.}
\vspace{-.2cm}
\label{fig:facade_3d_visuals}
\end{figure*}

Here, the task is to assign semantic label to every point in a point cloud and/or corresponding
multi-view 2D images.

\begin{figure}[!ht]
\captionsetup[subfigure]{labelformat=empty}
\begin{subfigure}{.155\textwidth}
\centering
\includegraphics[height=3.5cm]{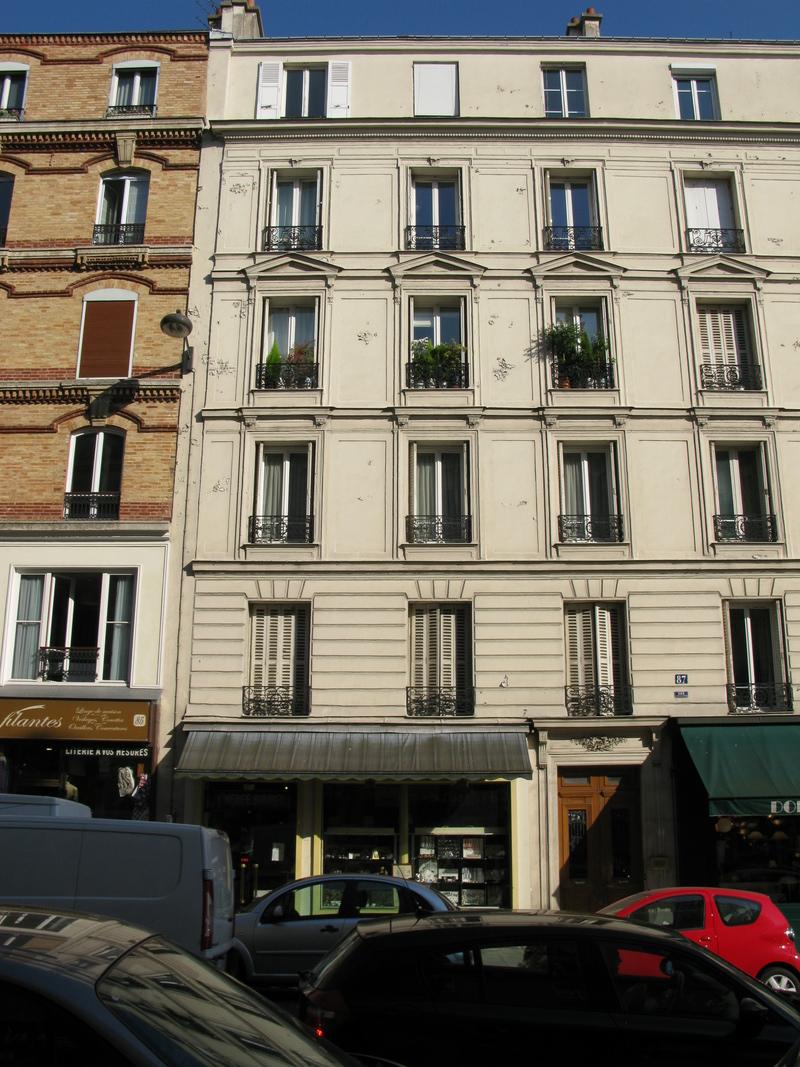}
\end{subfigure}
\begin{subfigure}{.155\textwidth}
\centering
\includegraphics[height=3.5cm]{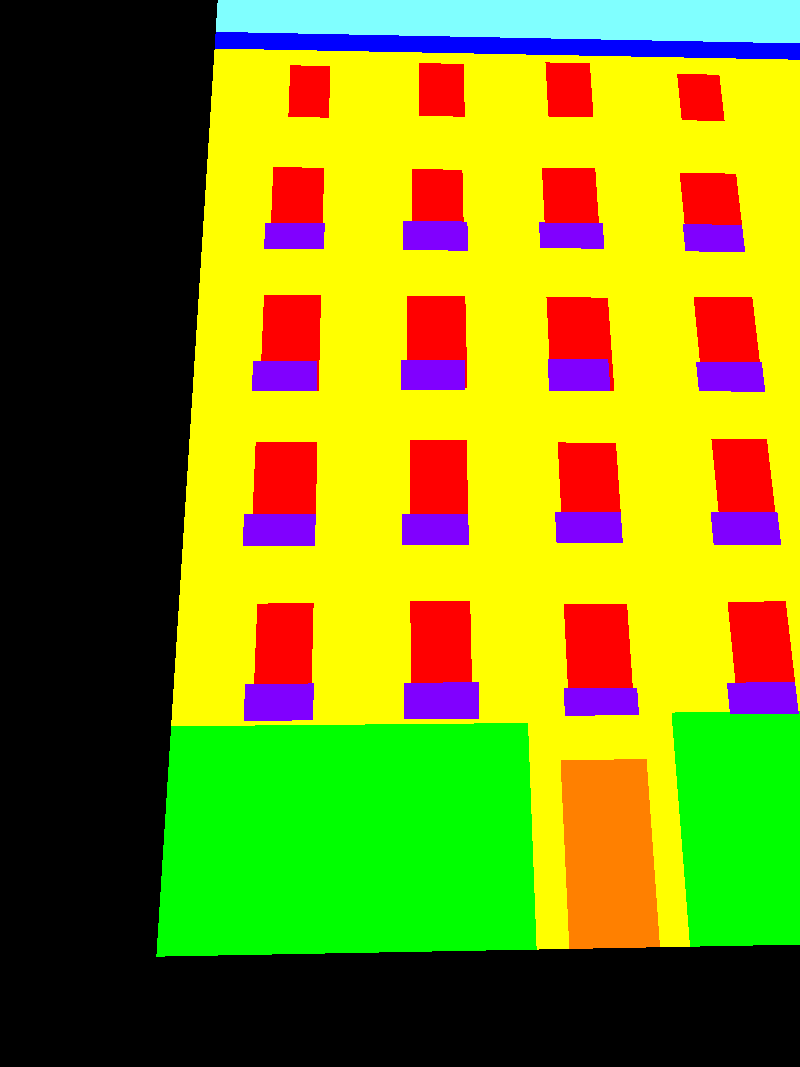}
\end{subfigure}
\begin{subfigure}{.155\textwidth}
\centering
\includegraphics[height=3.5cm]{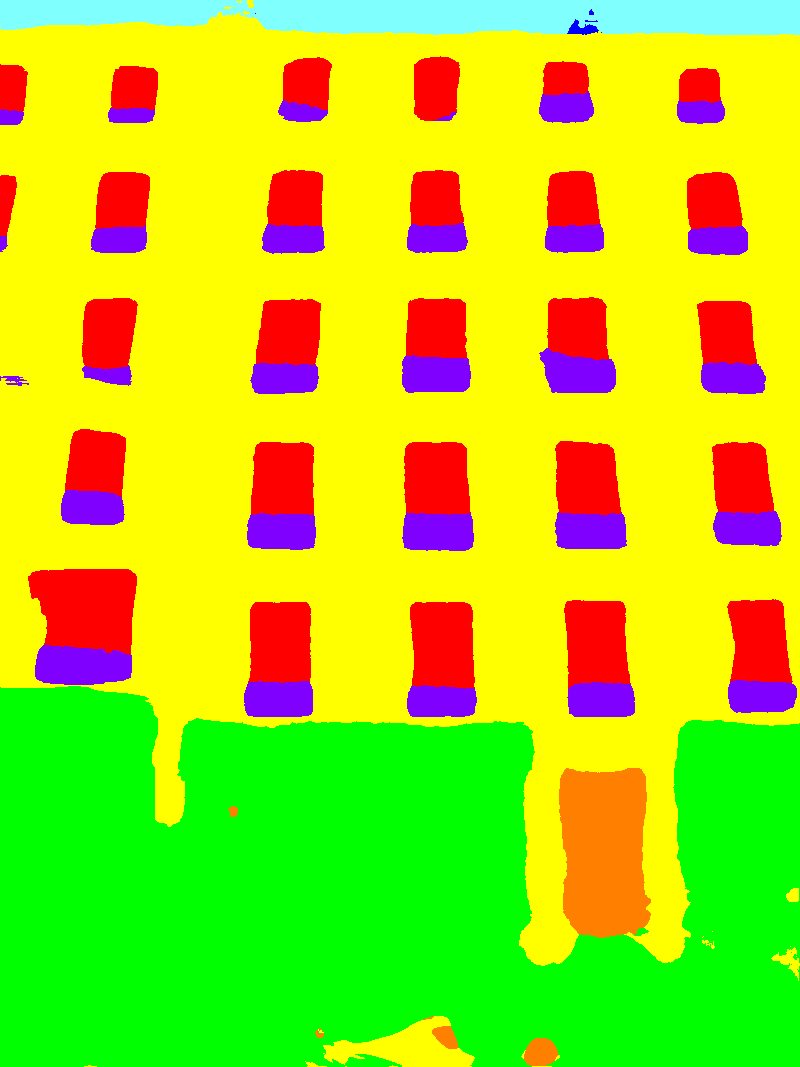}
\end{subfigure} \\[3pt]
\begin{subfigure}{.155\textwidth}
\centering
\includegraphics[height=3.5cm]{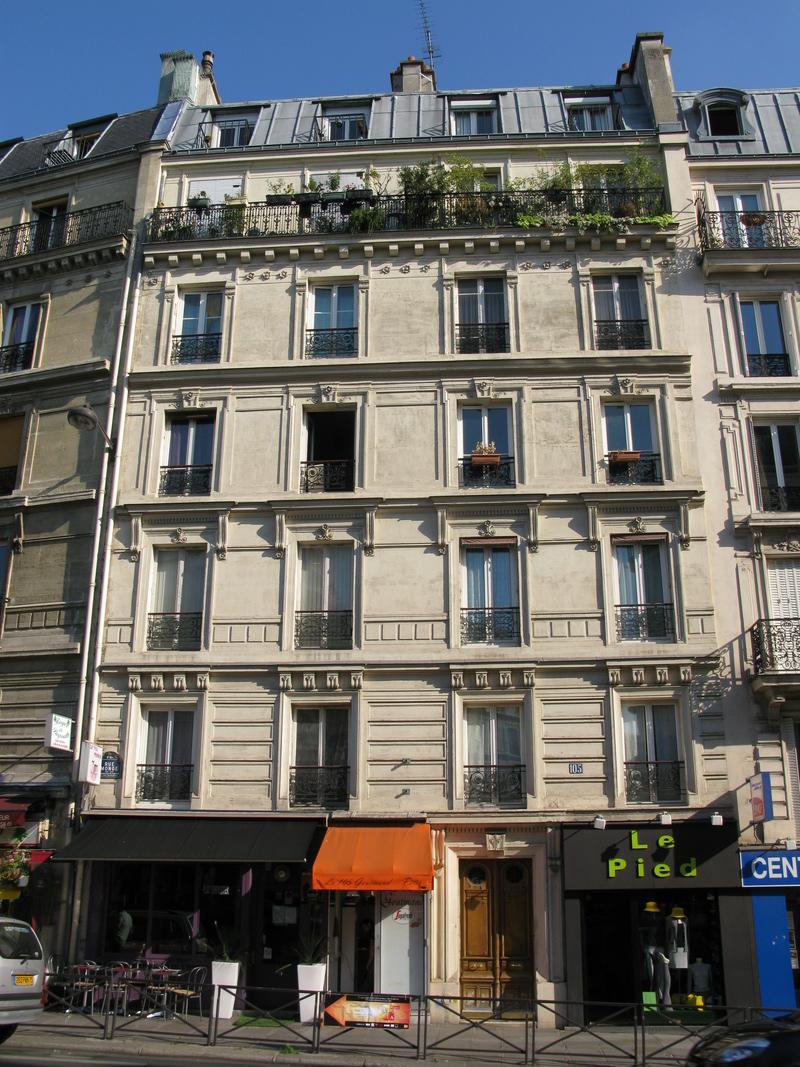}
\caption{Input}
\end{subfigure}
\begin{subfigure}{.155\textwidth}
\centering
\includegraphics[height=3.5cm]{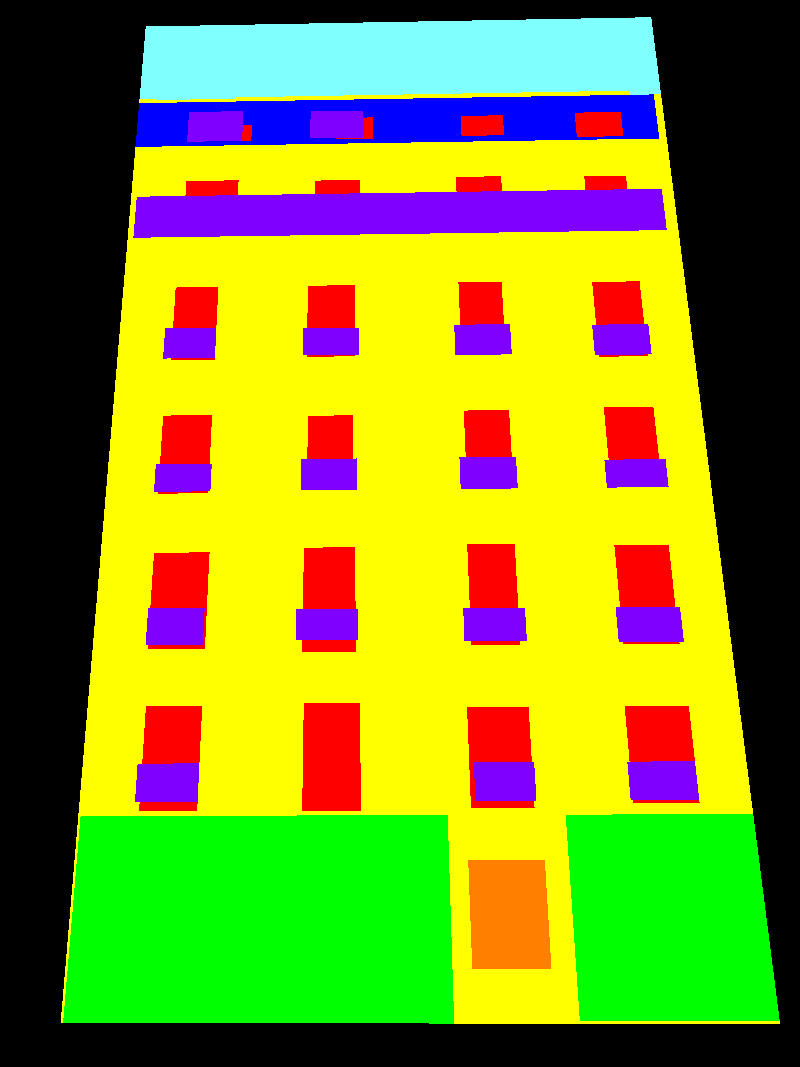}
\caption{Ground truth}
\end{subfigure}
\begin{subfigure}{.155\textwidth}
\centering
\includegraphics[height=3.5cm]{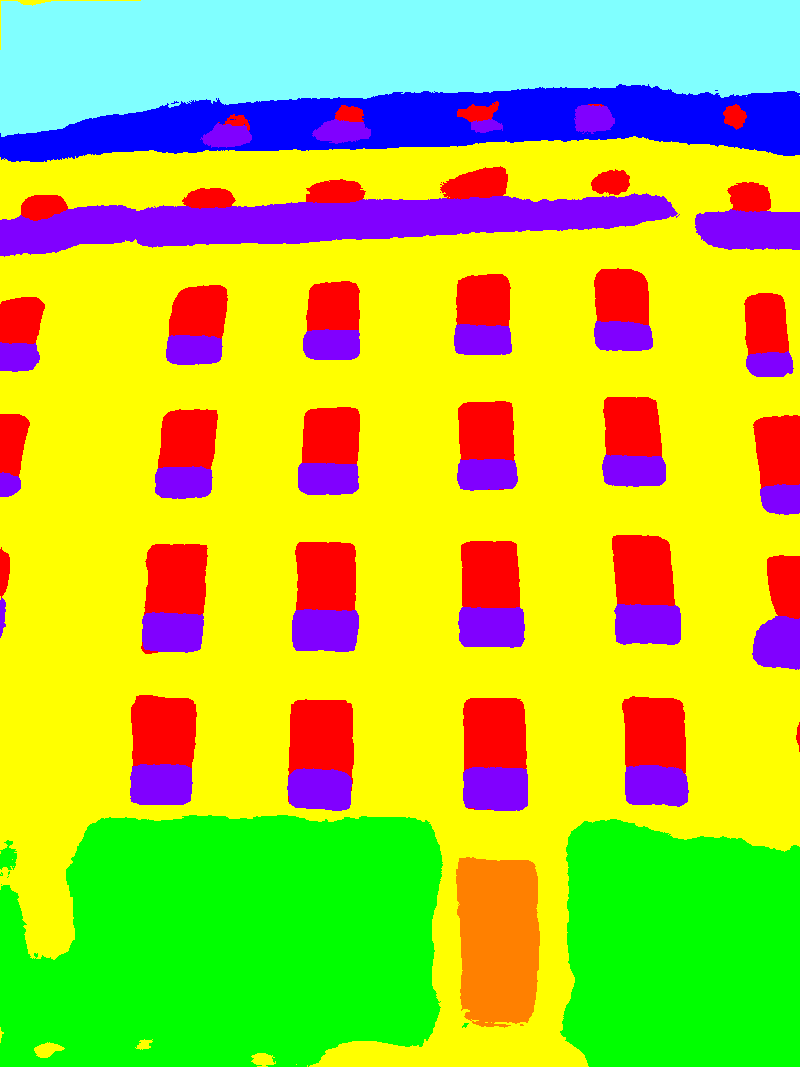}
\caption{\modeljoint}
\end{subfigure}
\vspace{-1mm}
\mycaption{2D facade segmentation}{Sample visual results of \modeljoint.}
\label{fig:facade_2d_visuals}
\vspace{-2mm}
\end{figure}

\vspace{-0.35cm}
\paragraph{Dataset.} RueMonge2014~\cite{riemenschneider2014learning} provides
a standard benchmark for 2D and 3D facade segmentation and also inverse procedural
modeling. The dataset consists of 428 high-resolution and multi-view images obtained
from a street in Paris. A point cloud with approximately 1M points is reconstructed
using the multi-view images. A ground-truth labeling with seven semantic classes of
door, shop, balcony, window, wall, sky and roof are provided for both 2D images and
the point cloud. Sample point cloud sections and 2D images with their corresponding ground truths
are shown in Figure~\ref{fig:facade_3d_visuals} and \ref{fig:facade_2d_visuals}
respectively. For evaluation, Intersection over Union (IoU) score is computed for
each of the seven classes and then averaged to get a single overall IoU.

\vspace{-0.35cm}
\paragraph{Point cloud labeling.} We use our \modelthree architecture for the task
of point cloud labeling on this dataset. We use 5 BCLs followed by a couple of 
$1\times1$ CONV layers. Input features to the network
comprise of a 7-dimensional vector at each point representing RGB color, normal and height above the ground. For all the BCLs, we use $XYZ$ lattice space ($L_{\text{3D}}$) with $\Lambda_0 = 64I_3$. Experimental results with average IoU and runtime are shown in 
Table~\ref{tab:facade_results_3d}. Results show that, with only 3D data, 
our method achieves an IoU of 65.4 which is a considerable improvement (6.2 IoU $\uparrow$) over the state-of-the-art deep network, OctNet~\cite{riegler2017octnet}.

Since this dataset comes with multi-view 2D images, one could leverage the
information present in 2D data for better point cloud labeling. 
We use \modeljoint to leverage 2D information and obtain better 3D segmentations. Table~\ref{tab:facade_results_3d} shows the experimental results when using
both the 2D and 3D data as input. \modeljoint obtains an average IoU of 69.8 outperforming
the previous state-of-the-art by a large margin (6.9 IoU $\uparrow$), thereby setting up a new
state-of-the-art on this dataset. This is also a significant improvement from the IoU obtained
with \modelthree demonstrating the benefit of leveraging 2D and 3D information in a joint
framework. Runtimes in Table~\ref{tab:facade_results_3d} also indicate that
our \model approach is much faster compared to traditional Autocontext techniques.
Sample visual results for 3D facade labeling are shown in Figure~\ref{fig:facade_3d_visuals}.

\vspace{-0.35cm}
\paragraph{Multi-view image labeling.} As illustrated in Section~\ref{sec:bpcn_2d3d}, we extend
2D CNNs with \modeljoint to obtain better multi-view image segmentation.
Table~\ref{tab:facade_results_2d} shows the results of multi-view image labeling on this dataset
using different techniques. Using DeepLab (CNN$_1$) already outperforms existing state-of-the-art
by a large margin. Leveraging 3D information via \modeljoint boosts
the performance to 70.6 IoU. An increase of 1.3 IoU from only using CNN$_1$ demonstrates
the potential of our joint 2D-3D framework in leveraging 3D information for better 2D segmentation.


\subsection{ShapeNet part segmentation}
\label{sec:exp_shapenet}

\begin{table*}[th!]
\scriptsize
    \mycaption{Results on ShapeNet part segmentation}{Class average mIoU, instance average mIoU and mIoU scores for all the categories on the task of point cloud labeling using different techniques.}
    \centering
    \small
\setlength{\tabcolsep}{2pt}
\begin{tabular}{ccccccccccccccccccc}
\toprule
\#instances                         &                                 &                                    & 2690      & 76   & 55   & 898  & 3758  & 69       & 787    & 392   & 1547 & 451    & 202        & 184  & 283    & 66     & 152        & 5271  \\ \hline
	\multicolumn{1}{l|}{}               & \multicolumn{1}{c|}{class} & \multicolumn{1}{c|}{instance} & \small{air-} & \small{bag}  & \small{cap}  & \small{car}  & \small{chair} & \small{ear-} & \small{guitar} & \small{knife} & \small{lamp} & \small{laptop} & \small{motor-} & \small{mug}  & \small{pistol} & \small{rocket} & \small{skate-} & \small{table} \\
	\multicolumn{1}{l|}{}               & \multicolumn{1}{c|}{avg.} & \multicolumn{1}{c|}{avg.} & \small{plane} & & & & & \small{phone} & & & & & \small{bike} & & & & \small{board} & \\
	\midrule
\multicolumn{1}{l|}{Yi \etal{~\cite{yi2016scalable}}} & \multicolumn{1}{c|}{79.0}     & \multicolumn{1}{c|}{81.4}         & 81.0      & 78.4 & 77.7 & 75.7 & 87.6  & 61.9     & 92.0   & 85.4  & 82.5 & 95.7   & 70.6     & 91.9 & \textbf{85.9}   & 53.1   & 69.8 & 75.3 \\
\multicolumn{1}{l|}{3DCNN{~\cite{qi2017pointnet}}}          & \multicolumn{1}{c|}{74.9}     & \multicolumn{1}{c|}{79.4}         & 75.1      & 72.8 & 73.3 & 70.0 & 87.2  & 63.5     & 88.4   & 79.6  & 74.4 & 93.9   & 58.7       & 91.8 & 76.4   & 51.2   & 65.3       & 77.1  \\
\multicolumn{1}{l|}{Kd-network{~\cite{klokov2017escape}}}         & \multicolumn{1}{c|}{77.4}     & \multicolumn{1}{c|}{82.3}         & 80.1      & 74.6 & 74.3 & 70.3 & 88.6  & 73.5     & 90.2   & \textbf{87.2}  & 81.0 & 94.9   & 57.4       & 86.7 & 78.1   & 51.8   & 69.9       & 80.3  \\
\multicolumn{1}{l|}{PointNet{~\cite{qi2017pointnet}}}       & \multicolumn{1}{c|}{80.4}     & \multicolumn{1}{c|}{83.7}         & \textbf{83.4}      & 78.7 & 82.5 & 74.9 & 89.6  & 73.0     & 91.5   & 85.9  & 80.8 & 95.3   & 65.2       & 93.0 & 81.2   & 57.9   & 72.8       & 80.6  \\
\multicolumn{1}{l|}{PointNet++{~\cite{qi2017pointnetpp}}}       & \multicolumn{1}{c|}{81.9}     & \multicolumn{1}{c|}{85.1}         & 82.4      & 79.0 & 87.7 & 77.3 & \textbf{90.8}  & 71.8     & 91.0   & 85.9  & 83.7 & 95.3   & 71.6       & 94.1 & 81.3   & 58.7   & 76.4       & \textbf{82.6}  \\
\multicolumn{1}{l|}{SyncSpecCNN{~\cite{yi2017syncspeccnn}}}       & \multicolumn{1}{c|}{82.0}     & \multicolumn{1}{c|}{84.7}         & 81.6 & 81.7 & 81.9 & 75.2 & 90.2 & 74.9 & \textbf{93.0} & 86.1 & \textbf{84.7} & 95.6 & 66.7 & 92.7 & 81.6 & 60.6 & \textbf{82.9} & 82.1  \\
\midrule
\multicolumn{1}{l|}{\modelthree}       & \multicolumn{1}{c|}{82.0}     & \multicolumn{1}{c|}{84.6}         & 81.9 & 83.9 & 88.6 & 79.5 & 90.1 & 73.5 & 91.3 & 84.7 & 84.5 & \textbf{96.3} & 69.7 & 95.0 & 81.7 & 59.2 & 70.4 & 81.3  \\
\multicolumn{1}{l|}{\modeljoint}       & \multicolumn{1}{c|}{\textbf{83.7}}     & \multicolumn{1}{c|}{\textbf{85.4}}         & 83.2 & \textbf{84.3} & \textbf{89.1} & \textbf{80.3} & 90.7 & \textbf{75.5} & 92.1 & 87.1 & 83.9 & \textbf{96.3} & \textbf{75.6} & \textbf{95.8} & 83.8 & \textbf{64.0} & 75.5 & 81.8  \\
\bottomrule
\end{tabular}
\label{tab:seg-results}
\end{table*}

The task of part segmentation is to assign a part category label to each point in a point cloud representing a 3D object.

\vspace{-0.35cm}
\paragraph{Dataset.}
The ShapeNet Part dataset~\cite{yi2016scalable} is a subset of ShapeNet, which contains 16681 objects from 16 categories, each with 2-6 part labels. The objects are consistently aligned and scaled to fit into a unit cube, and the ground-truth annotations are provided on sampled points on the shape surfaces. 
It is common to assume that the  category of the input 3D object is known, narrowing the possible part labels to the ones specific to the given object category.
We report standard IoU scores for evaluation of part segmentation. An IoU score
is computed for each object and then averaged within the objects in a category to compute mean IoU (mIoU)
for each object category. In addition to reporting mIoU score for each object category, we also report
`class average mIoU' which is the average mIoU across all object categories, and also `instance
average mIoU', which is the average mIoU across all objects.


\begin{figure}
\vspace{-1mm}
\begin{center}
\centerline{\includegraphics[width=\columnwidth]{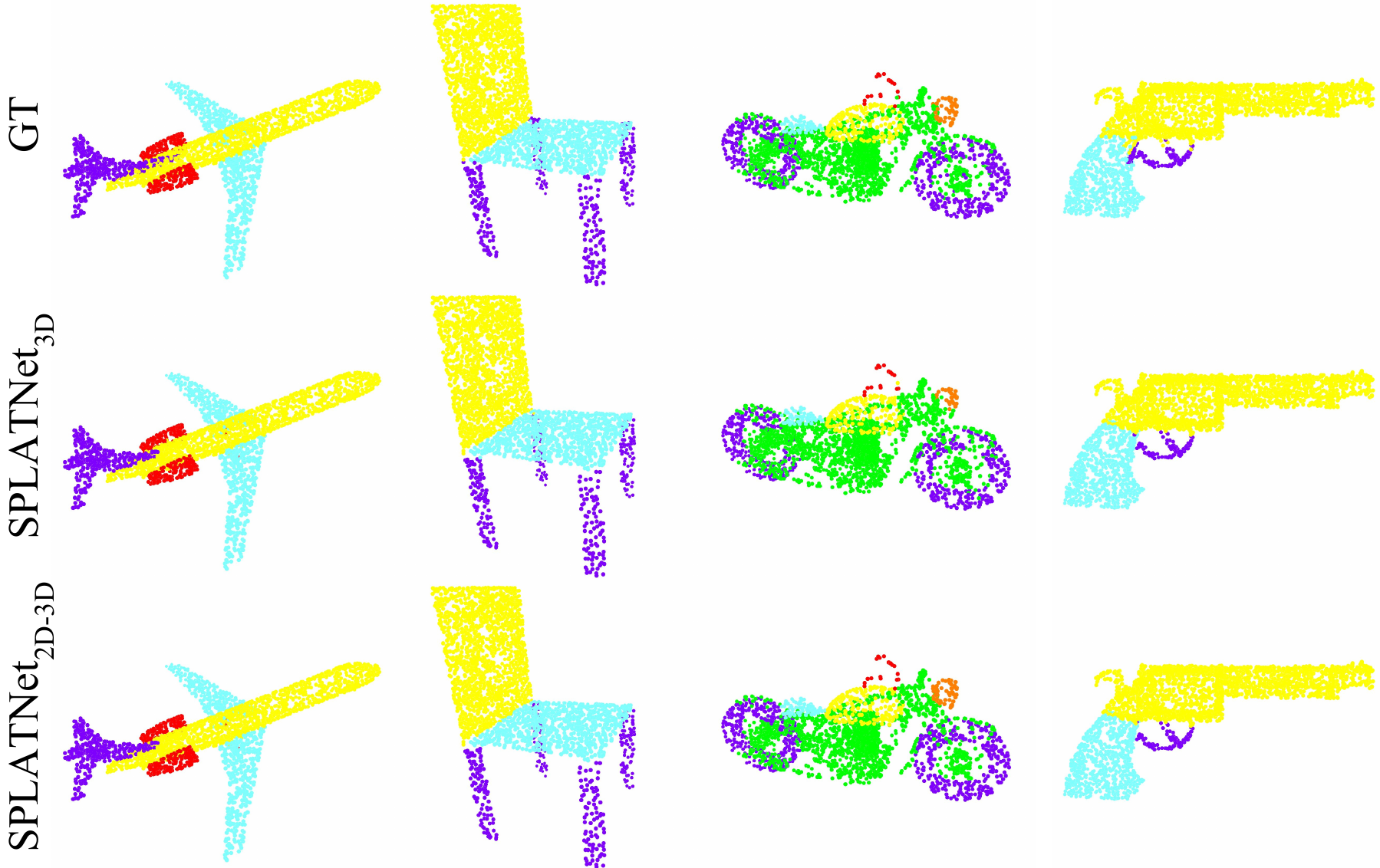}}
  \vspace{-0mm}
  \mycaption{ShapeNet part segmentation} {Sample visual results of \modelthree and \modeljoint.}
    \label{fig:shapenet_seg_samples}
\end{center}
\vspace{-8mm}
\end{figure}

\vspace{-0.35cm}
\paragraph{3D part segmentation.}
We evaluate both \modelthree and \modeljoint for this task. First, we discuss the architecture and results with \modelthree that uses only 3D point clouds as input.
Since the category of the input object is assumed to be known, we train separate networks
for each object category.
\modelthree network architecture for this taks is also composed of 5 BCLs.
Point locations $XYZ$ are used as input features as well as lattice features $L$ for all the BCLs
and the lattice scale for the first BCL layer is $\Lambda_0 = 64I_3$.
Experimental results are shown in Table~\ref{tab:seg-results}. \modelthree obtains a class average mIoU of 82.0 and an instance average mIoU of 84.6, which is on-par with the best networks that only take point clouds as input (PointNet++~\cite{qi2017pointnetpp} uses surface normals as additional inputs).


We also adopt our \modeljoint network, which operates on both 2D and 3D data, for this task.
For the joint framework to work, we need rendered 2D views and corresponding 3D locations for each pixel in the renderings. 
We first render 3-channel images: Phong shading~\cite{Phong:1975:ICG}, depth, and height from ground. Cameras are placed on the 20 vertices of a dodecahedron from a fixed distance, pointing towards the object's center. 
The 2D-3D correspondences can be generated by carrying the $XYZ$ coordinates of 3D points into the rendering rasterization pipeline so that each pixel also acquires coordinate values from the surface point projected onto it. Results in Table~\ref{tab:seg-results} show that incorporating 
2D information allows \modeljoint to improve noticeably from \modelthree with 1.7 and 0.8 increase in class and instance average mIoU respectively. 
\modeljoint obtains a class average IoU of 83.7 and an instance average IoU of 85.4, outperforming existing state-of-the-art approaches.

On one Nvidia GeForce GTX 1080 Ti, \modelthree{} runs at $9.4$ shapes$/$sec, while \modeljoint{} is slower at $0.4$ shapes$/$sec due to a relatively large 2D network
operating on 20 high-resolution ($512\times512$) views, which takes up more than $95\%$ of the computation time. In comparison, PointNet++ runs at $2.7$ shapes$/$sec on the same hardware\footnote{We use the public implementation released by the authors (\url{https://github.com/charlesq34/pointnet2}) with settings: $\text{model}=\text{`pointnet2\_part\_seg\_msg\_one\_hot'}$, $\text{VOTE\_NUM}=12$, $\text{num\_point}=3000$ (in consistence with our experiments).}.

\vspace{-0.35cm}
\paragraph{Six-dimensional filtering.} 
We experiment with a variant of \modelthree where an 
additional BCL with 6-dimensional position and normal lattice features ($XYZn_xn_yn_z$) is added between the last two $1\times1$ CONV layers. This modification gave only a marginal improvement of
$0.2$ IoU over standard \modelthree in terms of both class and instance average mIoU scores.


\section{Conclusion}
\label{sec:conclusion}

In this work, we propose the SPLATNet architecture for point cloud processing. SPLATNet directly takes point clouds as input and computes hierarchical and spatially-aware features with sparse and efficient lattice filters. In addition, SPLATNet allows an easy mapping of 2D information into 3D and vice-versa, resulting in a novel network architecture for joint processing of point clouds and multi-view images. Experiments on two different benchmark datasets show that the proposed networks compare favorably against state-of-the-art approaches for segmentation tasks. In the future, we would like to explore the use of additional input features (\eg, texture) and also the use of other high-dimensional lattice spaces in our networks.



\paragraph{Acknowledgements} Maji acknowledges support from NSF (Grant No. 1617917). Kalogerakis acknowledges support from NSF (Grant No. 1422441 and 1617333). Yang acknowledges support from NSF (Grant No. 1149783). We acknowledge the MassTech Collaborative grant for funding the UMass GPU cluster.


{\small
\bibliographystyle{ieee}
\bibliography{egbib}
}

\newpage
\clearpage
\appendix

{\raggedleft{} \bf \Large Supplementary}
\vspace{0.3cm}

\renewcommand\thesection{\Alph{section}}

In this supplementary material, we provide additional details and explanations to help readers gain 
a better understanding of our technique. 



\section{Point Cloud Density Normalization}

BCL has a normalization scheme to deal with uneven point density, or more specifically, the fact that some lattice vertices are supported by more data points than others.
Input signals are filtered directly with the learnable filter kernels, and are also filtered in a separate second round with their values replaced by $1$s with a Gaussian kernel. The filter responses in the second round are then used for normalizing responses from the first round.
This is similar
to using homogeneous coordinates, which are widely adopted in bilateral filtering implementations such as \cite{adams2010fast}.

\section{RueMonge2014 Facade Segmentation}

\paragraph{Network architecture of \modelthree.} We use 5 BCLs ($T=5$) 
followed by 
2 $1\times 1$ CONV layers in \modelthree for the facade segmentation task. 
We omit the initial $1\times 1$ CONV layer 
since we find it has no effect on the overall performance.
The number of output channels in each layer are: 
{\verb B64-B128-B128-B128-B64-C64-C7 }.
Note that although written as a linear structure, the network has skip connections from all BCLs 
(layers start with `{\verb B }') to the penultimate $1\times 1$ CONV layer. We use an initial scale 
$\Lambda_0=32I_3$ for scaling lattice features $XYZ$, and divide the scale in half after each BCL: 
$(32I_3, 16I_3, 8I_3, 4I_3, 2I_3)$. 
The unit of raw input features $XYZ$ is meter, with $Y$ 
(aligned with gravity axis) having a range of $7.1$ meters.
For all the BCLs, we use filters operating on one-ring neighborhoods on the lattice.

\vspace{-0.3cm}
\paragraph{Network architecture of \modeljoint.} We use \modelthree as described above as the 3D component 
of our 2D-3D joint model. The `2D-3D Fusion' component has 2 $1\times 1$ CONV layers: {\verb C64-C7 }. 
DeepLab~\cite{chen2014semantic} segmentation architecture is used as CNN$_1$. 
CNN$_2$ is a small network with 2 CONV layers: {\verb C32-C7 }, where the first layer has 
$3\times 3$ filters and 32 output channels, and the second one has $1\times 1$ filters and 7 output channels. 
We use $\Lambda_a=64$ and $\Lambda_b=1000$ 
for 2D$\leftrightarrow$3D projections with BCLs. Note that the dataset provides one-to-many mappings from 3D points 
to pixels. By using a very large scale (\ie, $\Lambda_b=1000$), 3D unaries are directly mapped to 
the corresponding 2D pixel locations without any interpolation. 

\vspace{-0.3cm}
\paragraph{Training.} We randomly sample facade segments of 60k points and use a batch size of 4 when 
training \modelthree. 
CNN$_1$ is initialized with Pascal VOC~\cite{Everingham15} pre-trained weights 
and fine-tuned for 2D 
facade segmentation.
Adam optimizer~\cite{kingma2014adam} with an initial learning rate 
of $0.0001$ is used for training both \modelthree and \modeljoint. 
Since the training data is small, we augment point cloud training data with random rotations, 
translations, and small color perturbations. We also augment 2D image data with small color 
perturbations during training.

\section{ShapeNet Part Segmentation}
\begin{figure*}[ht]
    \vspace{-1mm}
    \centering
    \begin{subfigure}[t]{.4\textwidth}
        \centering
        \includegraphics{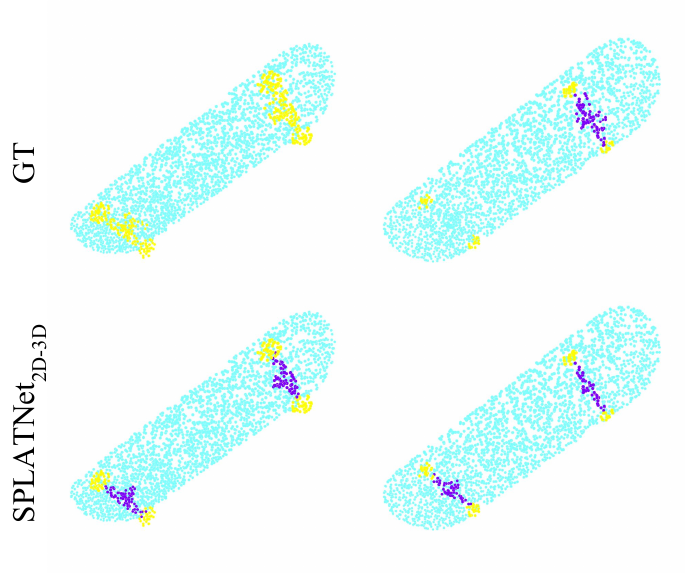}
        \caption{Incorrect labels}
    	\label{fig:errors_wrong}
    \end{subfigure}
    \begin{subfigure}[t]{.42\textwidth}
        \centering
        \includegraphics{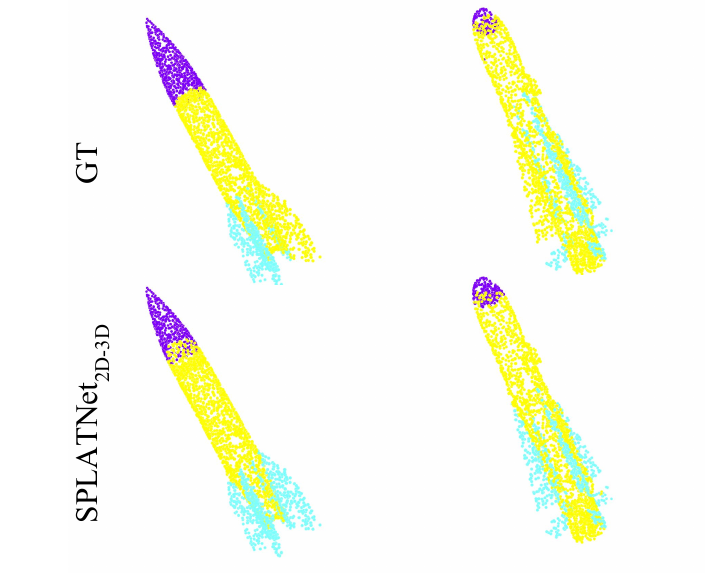}
        \caption{Incomplete labels}
    	\label{fig:errors_incomplete}
    \end{subfigure}\\
    \begin{subfigure}[t]{.4\textwidth}
        \centering
        \includegraphics{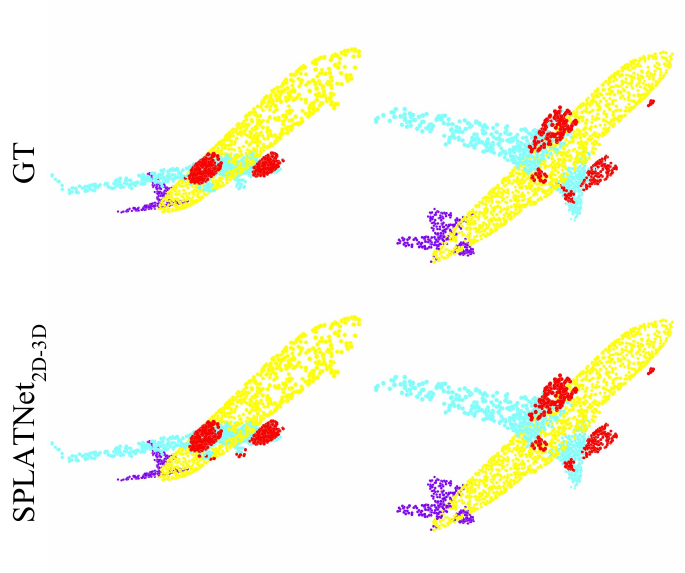}
        \caption{Inconsistent labels}
    	\label{fig:errors_inconsistent}
    \end{subfigure}
    \begin{subfigure}[t]{.42\textwidth}
        \centering
        \includegraphics{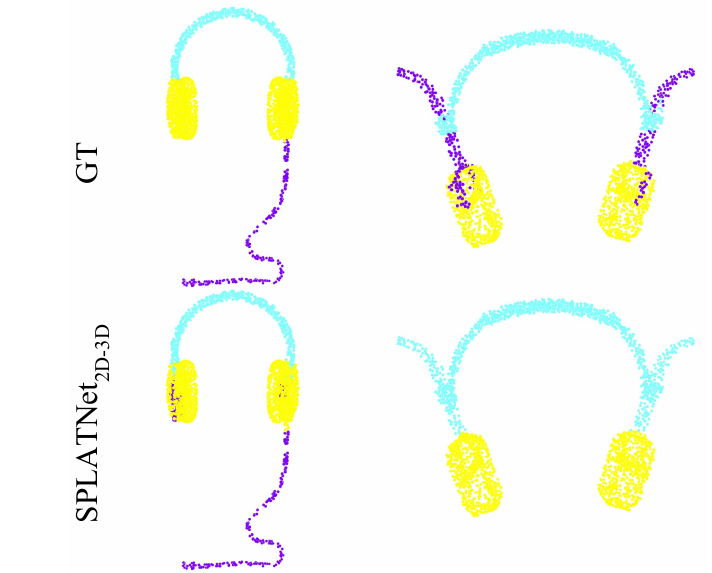}
        \caption{Confusing labels}
    	\label{fig:errors_confusing}
    \end{subfigure}
    \mycaption{Labeling issues in the ShapeNet Part dataset}{Four types of labeling issues are shown here. 
Two examples from the test set are given for each type, where the first row shows the ground-truth labels 
and the second row shows our predictions with \modeljoint. Our predictions 
appear to be
more accurate than the ground truth in some cases (see the skateboard axles in \ref{fig:errors_wrong} and the rocket fins in \ref{fig:errors_incomplete}).}
    \vspace{-1mm}
    \label{fig:errors}
\end{figure*}

\paragraph{Network architecture of \modelthree.} We use a $1\times 1$ CONV layer in the beginning,
followed by 5 BCLs ($T=5$), and then 2 $1\times 1$ CONV layers in 
\modelthree for the ShapeNet part segmentation task. 
The number of output channels in each layer are: 
{\verb C32-B64-B128-B256-B256-B256-C128-Cx }. 
`x' in the last CONV layer denotes the number of part categories, and 
ranges from 2-6 for different object categories.  
We use an initial scale 
$\Lambda_0=64I_3$ for scaling lattice features $XYZ$, and divide the scale in half after each BCL: 
$(64I_3, 32I_3, 16I_3, 8I_3, 4I_3)$. 

\vspace{-0.3cm}
\paragraph{Network architecture of \modeljoint.} We use \modelthree as described above as the 3D component 
of the joint model. The `2D-3D Fusion' component has 2 $1\times 1$ CONV layers: {\verb C128-Cx }. 
The same DeepLab architecture is used for CNN$_1$. 
We use $\Lambda_a=32$ in BCL$_{\text{2D}\rightarrow\text{3D}}$. 
Since 2D prediction is not needed, CNN$_2$ and 
BCL$_{\text{3D}\rightarrow\text{2D}}$ are omitted. 

\vspace{-0.3cm}
\paragraph{Training.} We train separate models for each object category. 
CNN$_1$ is initialized the same way as in the facade experiment. 
Adam optimizer with an initial learning rate 
of $0.0001$ is used. 
We augment point cloud data with random rotations, translations, and scalings during training. 

We train our networks until validation loss plateaus. Training \modelthree{} and \modeljoint{} take about $2.5$ and $3$ days respectively. 
With default settings, 
training PointNet++ takes $3.5$ days on the same hardware.

\vspace{-0.3cm}
\paragraph{Dataset labeling issues.}

We observed a few types of labeling issues in the ShapeNet Part dataset: 
\begin{itemize}
\vspace{-0.1cm}
\item Some object part categories are frequently labeled incorrectly. 
\Eg, skateboard axles are often mistakenly labeled as `deck' or `wheel' (Figure~\ref{fig:errors_wrong}). 
\vspace{-0.1cm}
\item Some object parts, \eg `fin' of some rockets, have incomplete range or coverage 
(Figure~\ref{fig:errors_incomplete}).
\vspace{-0.1cm}
\item Some object part categories are labeled inconsistently between shapes. \Eg, airplane landing gears 
are seen labeled as `body', `engine', or `wings' (Figure~\ref{fig:errors_inconsistent}). 
\vspace{-0.1cm}
\item Some categories have 
parts that are labeled as `other', which can be confusing for the classifier as these parts do not have
clear semantic meanings or structures. \Eg, in the case of earphones,
anything that is not `headband' or `earphone' are given the same label 
(`other')
(Figure~\ref{fig:errors_confusing}). 
\end{itemize}

The first two issues make evaluations and comparisons on the benchmark less reliable, while the other two  make 
learning ill-posed or unnecessarily hard for the networks. 

\end{document}